\documentclass{formatting/ssl/ssl}

\title{Shape of Memory: a Geometric Analysis of Machine Unlearning in Second-Order Optimizers}
\subtitle{Information is "learned" by changing the geometry of the second-order optimizer state. But what is the shape of the optimizer when forced to forget? We look at the geometry of machine unlearning.}
\author{Kennon Stewart\affilnum{1,2}}
\affiliation{\affilnum{1} Second Street Labs, Detroit, MI, USA\\
\affilnum{2} Department of Statistics, University of Michigan, Ann Arbor, MI, USA}
\corrauth{Kennon Stewart, Second Street Labs}
\email{kennon@secondstreetlabs.io}
\date{April 2026}
\version{v1}
\preprintnumber{2026-001}
\lablogo{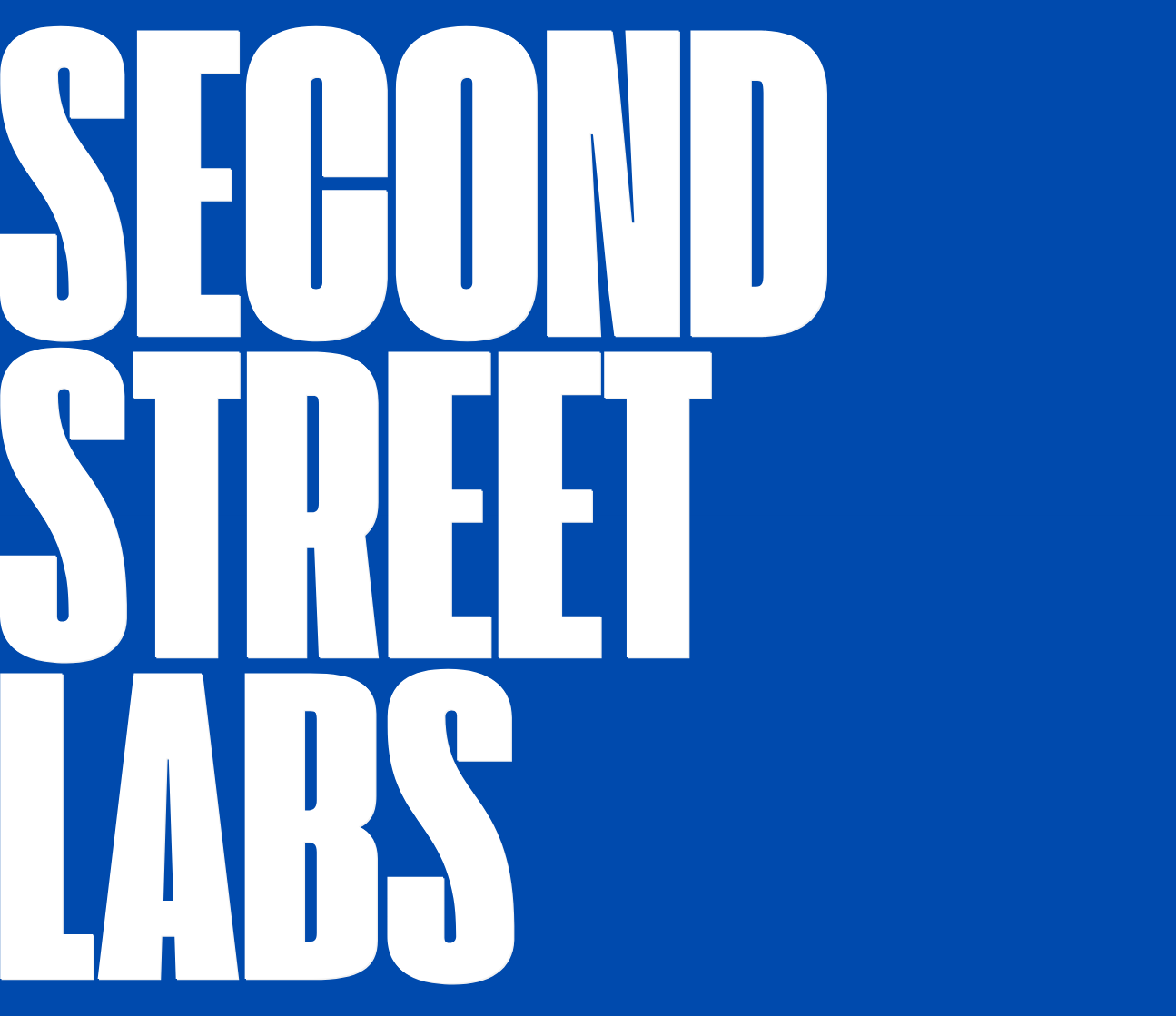}

\begin{document}

\begin{abstract}
We argue that current definitions of machine unlearning are underspecified for second-order optimizers. We compare first-order and second-order learners for their ability to handle the data deletion task with varying degrees of eigendecomposition to mimic the loss model memory. While both first and second-order methods realign with the ideal counterfactul in terms of performance and gradient, the second-order optimizer shows significant volatility in the optimizer state. This indicates residual information, supposedly deleted, that isn't detectable by first-order analysis. Various eigendecay treatments show that stability and information loss is regained only under controlled state pertubation where geometric information (or memory) is erased.
\end{abstract}
\maketitle

\section{Introduction}
Machine unlearning aims to remove the influence of specific data from a trained model, often to satisfy privacy, regulatory, or user-initiated deletion requirements. Existing work has largely focused on first-order optimization, where unlearning means that the post-deletion model is indistinguishable from the retrained ideal in terms of parameters. In this setting, the learner’s state is effectively summarized by its parameters, and deletion can be implemented with parameter perturbations \cite{Ginart_Guan_Valiant_Zou_2019}, partitioned retraining \cite{Bourtoule_Chandrasekaran_Choquette-Choo_Jia_Travers_Zhang_Lie_Papernot_2020}, or approximate inverse updates with certificates that bound residual leakage \cite{Sekhari_Acharya_Kamath_Suresh_2021, Ullah_Mai_Rao_Rossi_Arora_2021}.

Second-order optimizers make this difficult. They were designed to use curvature information to "quickly" gain information on the loss surface, but this additional state of information was never the focus of machine unlearning work.  For algorithms like Online Newton Step (ONS), this internal geometry defines a time-varying metric on parameter space, encoding how past data inform future predictions. A deletion request targets not only the parameters but also the optimizer’s memory, raising the question of what it means to "forget" data when the learning dynamics themselves depend on a compressed record of the entire trajectory.

This paper argues two things: that deleted information persists as volatility in the secon-order auxillary state, and that this volatility is reduced when the model undergoes geometric pertubation, mimicking the erasure of second-order information.

Current formulations of certified machine unlearning are underspecified for stateful optimizers. Definitions that equate successful unlearning with parameter similarity ignore the possibility that higher-order states retain information about deleted observations, even when external performance appears unaffected. This complicates data privacy work that seeks to efficiently remove the data from a model while only measuring performance and parameter difference.

We do not attempt to solve the problem of certified machine unlearning. We instead pose the problem as under-specified, and pose the challenge as a task for future research.

\section{Related Work}

There is a substantial body of work on convex optimization, with applications in online learning, data deletion, and machine unlearning. We review the pieces most relevant to our work while acknowledging the breadth of research available.

\subsection{Online Convex Optimization.} 
Convex optimization methods aim to efficiently locate a global optimum of a convex objective, subject to user-specified constraints. In the online setting, observations are revealed incrementally to the learner, requiring the model to update its parameters sequentially rather than in batches. This is formalized in the online convex optimization problem, where performance is measured relative to some ideal comparator \cite{Boyd_Vandenberghe_2023}.

First-order methods such as gradient descent remain popular due to their simplicity and minimal memory requirements. Each update consists of observing a gradient, selecting a step size, and applying a recursive update rule. However, these methods discard nearly all historical information beyond the current iterate.

Second-order methods, including Newton-style algorithms, trade additional memory and computation for faster convergence and improved conditioning. These methods explicitly retain curvature information, raising a fundamental design question: how much information from the optimization trajectory must be stored to ensure robust and efficient learning?

\subsection{Machine Unlearning.} 

The problem of data deletion was formalized rigorously in previous works.~\cite{Dwork_Roth_2014, Ullah_Mai_Rao_Rossi_Arora_2021, Ginart_Guan_Valiant_Zou_2019}, who propose proactive mechanisms that render learned models insensitive to the inclusion or exclusion of any single training point. Under this framework, no explicit deletion operation is required post hoc, as privacy is guaranteed by construction.

Subsequent work explored practical implementations of proactive unlearning. Bourtoule et al.~\cite{Bourtoule_Chandrasekaran_Choquette-Choo_Jia_Travers_Zhang_Lie_Papernot_2020} propose partitioned training, where each model component is trained on a disjoint subset of the data and retrained upon deletion. While effective, these approaches incur substantial performance and retraining costs.

Later work relaxes the requirement of perfect privacy in favor of approximate deletion guarantees. Ginart et al. frame unlearning as decremental learning, measuring success via distance in parameter space from an ideal model that never observed the deleted point \cite{Ginart_Guan_Valiant_Zou_2019}. Guo et al. propose a Newton-based deletion step that removes the influence of a data point in a single update \cite{Guo_Goldstein_Hannun_Maaten_2023}, with any residual leakage mitigated through calibrated noise \cite{Chen_Zhang_Wang_Backes_Humbert_Zhang_2021}. These approximate guarantees significantly improve practicality but define unlearning narrowly as parameter matching to a counterfactual model.

\subsection{Unlearning via Second-Order Approximations.} 

Second-order methods are particularly attractive for unlearning due to their explicit modeling of curvature. Classical Newton methods require storing and inverting the Hessian, which scales quadratically with parameter dimension. Quasi-Newton methods such as BFGS alleviate this cost by maintaining low-rank approximations to the inverse Hessian.

Nocedal and Wright formalized these techniques and demonstrated that curvature information can substitute for repeated gradient evaluations \cite{Byrd_Lu_Nocedal_Zhu_1995}. In the online setting, Mokhtari et al. proposed an online L-BFGS algorithm that retains a constant number of curvature pairs while achieving global convergence guarantees \cite{Mokhtari_Ribeiro_2014}. These methods emphasize efficiency and scalability but are typically analyzed under the assumption of monotonic data accumulation rather than adversarial deletions.

\subsection{Online and Continuous Learning} 

Recent work has emphasized the role of \emph{stateful optimizers}, which reason over some compressed form of information, called a iterate. This iterate can be a preconditioner matrix, curvature estimates, or any sufficient information state (more on this in future work). These methods address limitations of memoryless algorithms, particularly in nonstationary environments, and have been shown to achieve improved convergence rates and stability in online learning settings.

However, existing analyses of sequential learning largely assume that the optimization trajectory evolves monotonically as new data arrive. Deletions, when considered at all, are typically treated as retraining events or as perturbations to the model parameters. This perspective neglects the fact that stateful optimizers accumulate higher-order information that is itself shaped by the full training history.

While much prior work evaluates unlearning by comparing parameter vectors to a counterfactual model, this criterion is insufficient for stateful learners whose internal geometry encodes additional information. Sequential deletions can induce persistent discrepancies between the realized optimizer state and its counterfactual counterpart, even when parameter estimates appear similar. Our work departs from parameter-centric analyses by explicitly studying the evolution, distortion, and hysteresis of optimizer state under online deletions.

\section{Problem Formulation}

We study online convex optimization under adversarial deletions. Our learner progresses through discrete episodes $t = 1,2,\dots,T$. Let $\mathcal W \subset \mathbb R^d$ be a convex parameter domain. At each round $t$, the learner selects parameters $w_t \in \mathcal W$, observes a convex loss function $\ell_t$, and incurs loss $\ell_t(w_t)$.

\subsection{Learner State and Update Rules}

We explicitly distinguish between the learner’s parameters and its internal optimizer state. At time $t$, the learner state is
\[
S_t = (w_t, \mathcal A_t),
\]
where $w_t$ denotes the model parameters and $\mathcal A_t$ denotes an auxiliary state maintained by the optimization algorithm.

For first-order methods such as Online Gradient Descent (OGD), the second-order auxiliary state is empty. In contrast, second-order methods maintain a nontrivial state that aggregates historical information and influences future updates.

\subsection{Online Gradient Descent (OGD)}

As a first-order baseline, we consider Online Gradient Descent \cite{Cesa-Bianchi_Lugosi_2006}. Given step size $\eta > 0$, the update rule is
\[
w_{t+1} = w_t - \eta \nabla \ell_t(w_t),
\quad g_t = \nabla \ell_t(w_t).
\]
OGD retains no second-order or historical information beyond the current iterate. As a result, deletions affect only the parameter vector and not any accumulated optimizer state.

\subsection{Online Newton Step (ONS)}

For second-order learning, we study Online Newton Step (ONS). ONS maintains a positive definite matrix
\[
A_t = \lambda I + \sum_{s=1}^t g_s g_s^\top,
\]
where $\lambda > 0$ ensures numerical stability. The matrix $A_t$ defines a time-varying inner product
\[
\|x\|_{A_t}^2 = x^\top A_t x,
\]
and the parameter update is
\[
w_{t+1} = w_t - \eta A_t^{-1} g_t.
\]
Geometrically, ONS performs steepest descent with respect to the metric induced by $A_t$. The matrix $A_t$ constitutes an internal optimizer state that encodes accumulated curvature and directional sensitivity.

It's worth noting that the Online Newton Step is invariant to affine changes in the loss surface. Linear differences between the counterfactual and unlearned model will not change the trajectory of the descent. 
\subsection{Deletion Model}

At specified times $\tau$, an adversary issues deletion requests for previously observed examples. A deletion removes the influence of a past gradient $g_\tau$ from the learning process. Importantly, deletions may act on both components of the learner's internal state,
\[
\mathcal{S}_t = (w_t, A_t),
\]
where $w_t \in \mathbb{R}^d$ represents the model parameters as well as the second-order iterate at time $t,$ $A_t \in \mathbb{R}^{d \times d}$. We do not assume that the learner has access to the full training history or the computational budget to recompute $A_t$ from scratch.

\section{Methods: Second-Order State Intervention}
\subsection{Experimental Setup}

We study online deletion in a controlled linear convex learning problem using two learners: Online Gradient Descent (OGD) as a first-order baseline and Online Newton Step (ONS) as a second-order stateful optimizer. Each experiment is run over a horizon of $T=400$ rounds in dimension $d=2$. A deletion event is issued at time $\tau = 200$, removing the most recent $10$ observations prior to deletion. We evaluate both stationary and drifting data streams, with the drifting setting used to separate ordinary adaptation to nonstationarity from deletion-induced distortion in optimizer state. All reported results are aggregated over $20$ random seeds.

For OGD, we use an initial step size $\eta_0 = 0.6$ with a square-root decay schedule and projection radius $R = 5.0$. For ONS, we use regularization parameter $\delta = 1.0$, learning-rate parameter $\eta = 1.0$, and the same radius constraint $R = 5.0$. In both cases, we compare the realized post-deletion trajectory to a counterfactual trajectory trained on the same stream with the deleted observations removed. This design allows us to measure how closely the learner recovers the path it would have followed had the deleted data never been observed.

For second-order interventions, we apply deletion-time perturbations directly to the ONS state matrix at time $\tau$. We consider two spectral treatments. Partial Reset subtracts a constant $\alpha$ from the eigenvalues of the optimizer state, with positivity thresholding, using $\alpha \in \{0.3, 0.5, 0.7\}$. Eigenvalue Decay rescales the spectrum by $\beta \in \{0.5, 0.7, 0.9\}$. We also include a baseline condition with standard deletion but no explicit spectral intervention. These settings are chosen to span light, moderate, and aggressive distortions of stored second-order information while preserving numerical stability.

We report cumulative regret, tracking error, instantaneous parameter shock at deletion time, and state-space diagnostics for ONS, including trace, condition number, and cosine alignment of update directions or aggregate geometry. 

\subsection{Comparators and Metrics}
We report the regret with respect to the best fixed parameter in hindsight over the horizon $T$:
\[
R_T = \sum_{t=1}^T f_t(w_t) - \min_{w \in \mathcal{W}} \sum_{t=1}^T f_t(w)
\]
We also measure tracking error, which is the Euclidean distance between the observed and counterfactual models in terms of parameter space:
\[
E_t = \|w_t - w_t^{(-U)}\|_2
\]
and the instantaneous parameter shock is instantaneous deviation in parameter space precisely at the deletion time $\tau$:
\[
\Delta_\tau = \|w_{\tau+1} - w_{\tau+1}^{(-U)}\|_2
\]
Finally, we analyze the spectral properties of the trace, eigenvalue distribution, and condition number of second-order state $A_t$, and the eigenvalue distributions characterize how unlearning interventions permanently alter internal curvature and learning geometry.

\begin{figurepair}
  \figurepanel{\includegraphics[width=\linewidth]{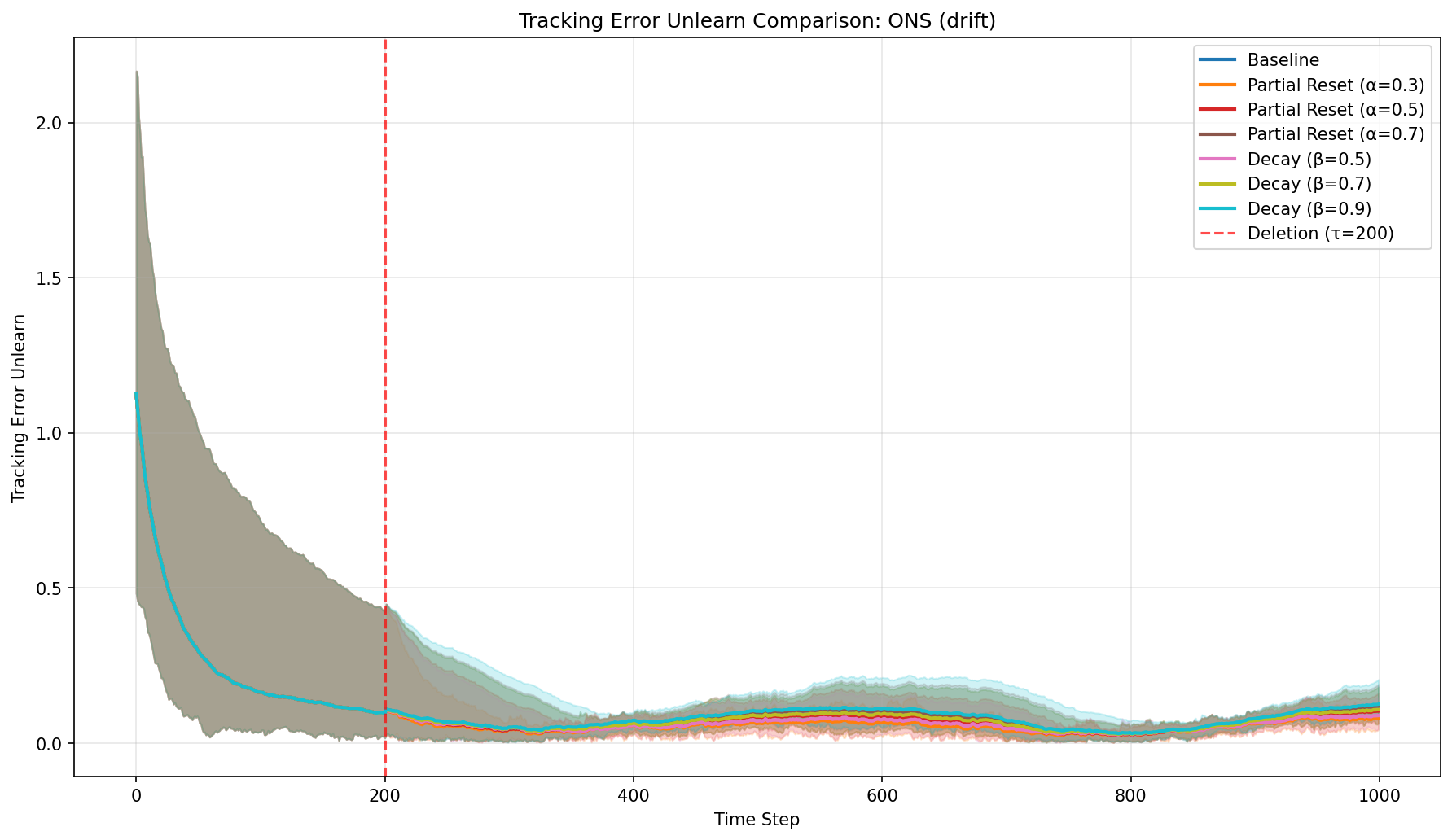}}{The figure shows the tracking error of the optimizer at different points of the learning process for 20 seeds in the drifting data-generating process. The tracking error converges to 0 for the first ~200 iterations of the simulation. Following the initial collapse to 0, there remains some additional error.}
  \hfill
  \figurepanel{\includegraphics[width=\linewidth]{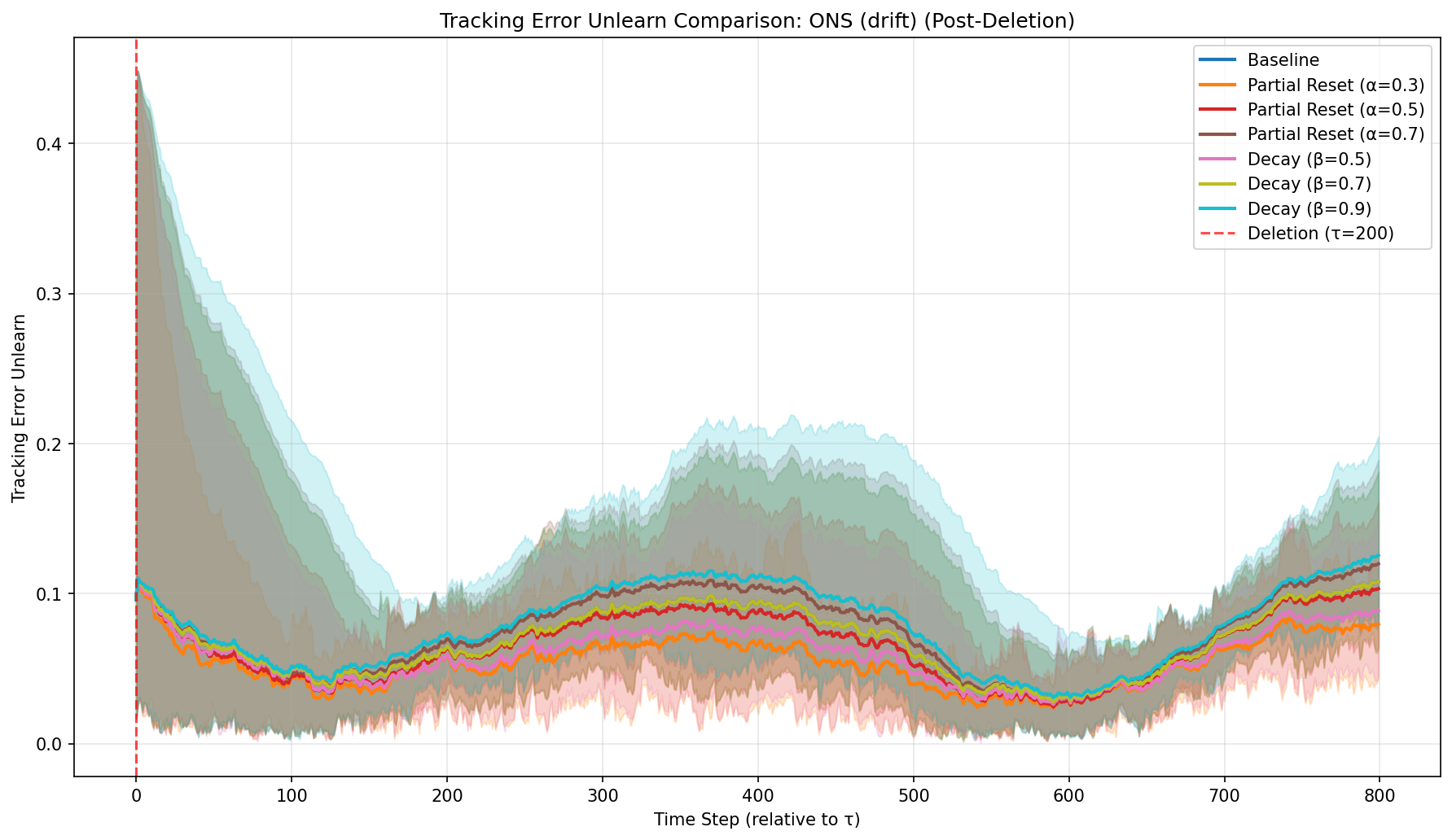}}{The figure shows the tracking error specifically following the deletion event. The trajectories diverge following the deletion event, with the degree of divergence increasing with the magnitude of the intervention.}
  \caption{Tracking error for second-order optimizer states under various spectral treatments. Figure (a) shows the simulation timeline in its entirety, while (b) specifically shows the tracking error from deletion at time $\tau=200$ to $t=1000.$}
  \label{fig:tracking-errors}
\end{figurepair}

\section{Results}

We compare first-order and second-order online learners under deletion-time interventions to evaluate whether conventional performance metrics are sufficient to characterize unlearning. Across experiments, three patterns recur. First, OGD exhibits finite-time recovery after deletion, with recovery substantially slower in drifting environments than in stationary ones. Second, ONS often returns quickly to its pre-deletion regret scale even when its internal state remains visibly misaligned with the counterfactual trajectory. Third, spectral interventions applied to the ONS state alter optimizer geometry far more strongly than they alter external performance, suggesting that deletion in second-order learners is primarily a problem of state alignment rather than immediate predictive failure.

\begin{table}[t]
\centering
\caption{Deletion-response summary across learners and intervention settings. Entries are mean $\pm$ standard deviation over 20 random seeds. Recovery time is measured in rounds after deletion. Overshoot denotes the maximum post-deletion deviation relative to the pre-deletion reference level, and parameter shock denotes the instantaneous displacement at deletion time.}
\label{tab:deletion-summary}
\small
\begin{tabular}{lllccc}
\toprule
Model & Environment & Intervention & Recovery Time & Overshoot & Param. Shock \\
\midrule
OGD & Stationary & Baseline & $22.75 \pm 59.04$ & $0.025 \pm 0.005$ & $0.014 \pm 0.008$ \\
OGD & Drifting   & Baseline & $78.35 \pm 18.65$ & $0.734 \pm 0.102$ & $0.049 \pm 0.010$ \\
\midrule
ONS & Drifting & Baseline               & $0.05 \pm 0.22$ & $1.35 \pm 6.22$ & $0.065 \pm 0.030$ \\
ONS & Drifting & Partial Reset ($\alpha=0.3$) & $0.05 \pm 0.22$ & $1.03 \pm 4.80$ & $0.065 \pm 0.030$ \\
ONS & Drifting & Partial Reset ($\alpha=0.5$) & $0.05 \pm 0.22$ & $1.24 \pm 5.72$ & $0.065 \pm 0.030$ \\
ONS & Drifting & Partial Reset ($\alpha=0.7$) & $0.05 \pm 0.22$ & $1.35 \pm 6.22$ & $0.065 \pm 0.030$ \\
ONS & Drifting & Decay ($\beta=0.5$)   & $0.05 \pm 0.22$ & $1.24 \pm 5.71$ & $0.065 \pm 0.030$ \\
ONS & Drifting & Decay ($\beta=0.7$)   & $0.05 \pm 0.22$ & $1.35 \pm 6.22$ & $0.065 \pm 0.030$ \\
ONS & Drifting & Decay ($\beta=0.9$)   & $0.05 \pm 0.22$ & $1.43 \pm 6.54$ & $0.065 \pm 0.030$ \\
\bottomrule
\end{tabular}
\end{table}

\begin{figurepair}
  \figurepanel{\includegraphics[width=\linewidth]{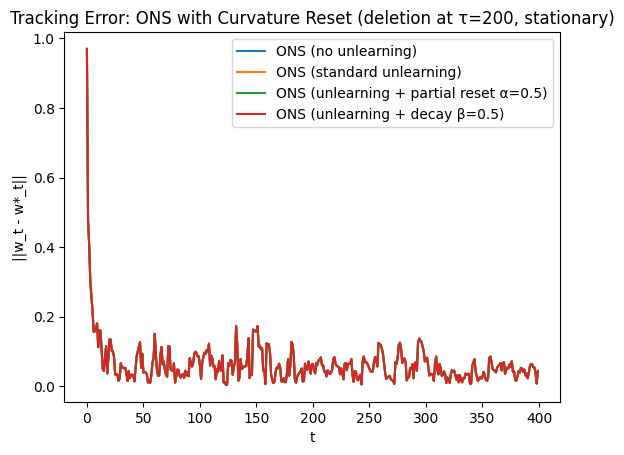}}{The figure shows the tracking error with respect to the ideal counterfactual model. A larger tracking error corresponds to a high difference in parameter values, which indicates some difference in model trajectories. We see a negligible difference in parameter values, indicating that the models retain similar trajectories pre- and post-deletions.}
  \hfill
  \figurepanel{\includegraphics[width=\linewidth]{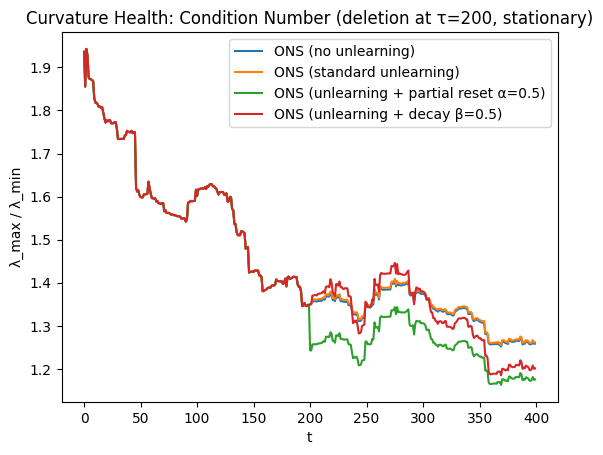}}{The figure shows the condition number of the second-order iterate. The difference between the first and second eigenvalues is a measure of the model's sensitivity to small changes in parameter space. For stationary distributions there is no curvature, and so the condition number predictably decays to 1. But once the intervention is applied, the partial reset drops exactly $\alpha=0.5,$ while the decay intervention notably increases the sensitivity of the iterate (ie. the stepwise changes in the trajectory are scaled up in the same direction).}
  \caption{Tracking error and condition number measure the difference in parameter between the observed and counterfactual models.}
  \label{fig:curvature-health}
\end{figurepair}

\subsection{Online Gradient Descent Recovers Counterfactual Performance.}

We begin with OGD as a first-order baseline. In the stationary setting, deletion only creates a small transient disturbance: the mean parameter shock is $0.0138 \pm 0.0079$, the mean overshoot is $0.0255 \pm 0.0053$, and the mean recovery time is $22.75$ rounds. In the drifting setting, the same deletion produces a larger disruption, with mean parameter shock $0.0488 \pm 0.0100$, mean overshoot $0.7337 \pm 0.1021$, and mean recovery time $78.35 \pm 18.65$ rounds. These results indicate that for first-order learners, deletion itself is not the sole determinant of recovery behavior. The ambient difficulty of tracking a moving comparator substantially lengthens the return to pre-deletion performance.

In OGD, the learner retains no auxiliary curvature state, so deletion affects only the parameter trajectory. The finite recovery times therefore provide a useful baseline for what ordinary post-deletion adaptation looks like in the absence of persistent optimizer memory.

\subsection{Online Newton Step Recovers First-Order Information.}

The ONS experiments under drift reveal a different phenomenon. Across all intervention settings, the regret-based recovery time is essentially immediate, with mean recovery time $0.05$ rounds for every treatment. On its face, this would suggest that ONS is highly robust to deletion. However, this interpretation is incomplete. The same runs exhibit nontrivial parameter shock at deletion, with mean parameter shock approximately $0.0654 \pm 0.0303$ across interventions, and substantial post-deletion overshoot ranging from $1.03$ to $1.43$.

Taken together, these numbers show that scalar performance summaries can return to their nominal scale even while the post-deletion trajectory has been displaced. In other words, ONS appears to recover quickly when viewed through regret alone, but this rapid recovery does not certify agreement with the counterfactual state. This is consistent with the visual diagnostics in Figure~\ref{fig:tracking-errors}, where post-deletion trajectories remain separated despite broadly similar large-scale performance.

\subsection{Stronger Hessian Suppression Enables Stronger Recovery.}

Among the ONS interventions, mild partial reset is the least disruptive in the current drifted experiments. The $\alpha=0.3$ partial reset achieves the smallest mean overshoot ($1.026$) and the lowest final regret ($5.402$) among the reported treatments. At the other end of the spectrum, aggressive decay with $\beta=0.9$ produces the largest overshoot ($1.425$) and the largest final regret ($5.747$). Intermediate interventions fall between these extremes, with only modest separation in final regret but clearer differences in post-deletion trajectory behavior.

This pattern suggests that stronger suppression of stored curvature does not reconstruct the counterfactual geometry. Instead, aggressive interventions tend to introduce a larger geometric discontinuity while leaving the learner broadly functional. The effect is therefore not best described as simple performance degradation. Rather, the interventions modify the metric under which ONS continues to adapt, changing the path of optimization without causing catastrophic forgetting.

The mean parameter shock is effectively identical across ONS intervention settings. The immediate displacement at deletion is therefore not the main source of separation between treatments. The differences emerge later, during the post-deletion evolution of the learner under a modified second-order state. This is expected if the interventions act primarily by rewriting optimizer geometry rather than by an individual first-order perturbation.

\subsection{Volatility Persists in the Second-Order Iterate.}

\begin{figurepair}
  \figurepanel{\includegraphics[width=\linewidth]{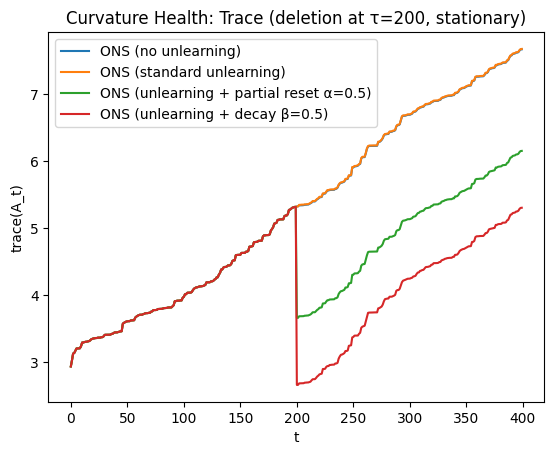}}{The figure shows the condition number of the Newton Step preconditioner $A_t$. At the time of intervention, the condition number $\kappa$ predictably degrades for the treatment groups to the appropriate degree. The partial reset intervention is equivalent to a scalar subtraction operation on the second-order iterate. We subtract $\alpha$ from the values to reduce the degree of curvature/shape in a brute-force sense. The decay intervention is more subtle. It scales the eigenvalues by some constant value, reducing the variance of the eigenvalues in addition to their magnitudes.}
  \hfill
  \figurepanel{\includegraphics[width=\linewidth]{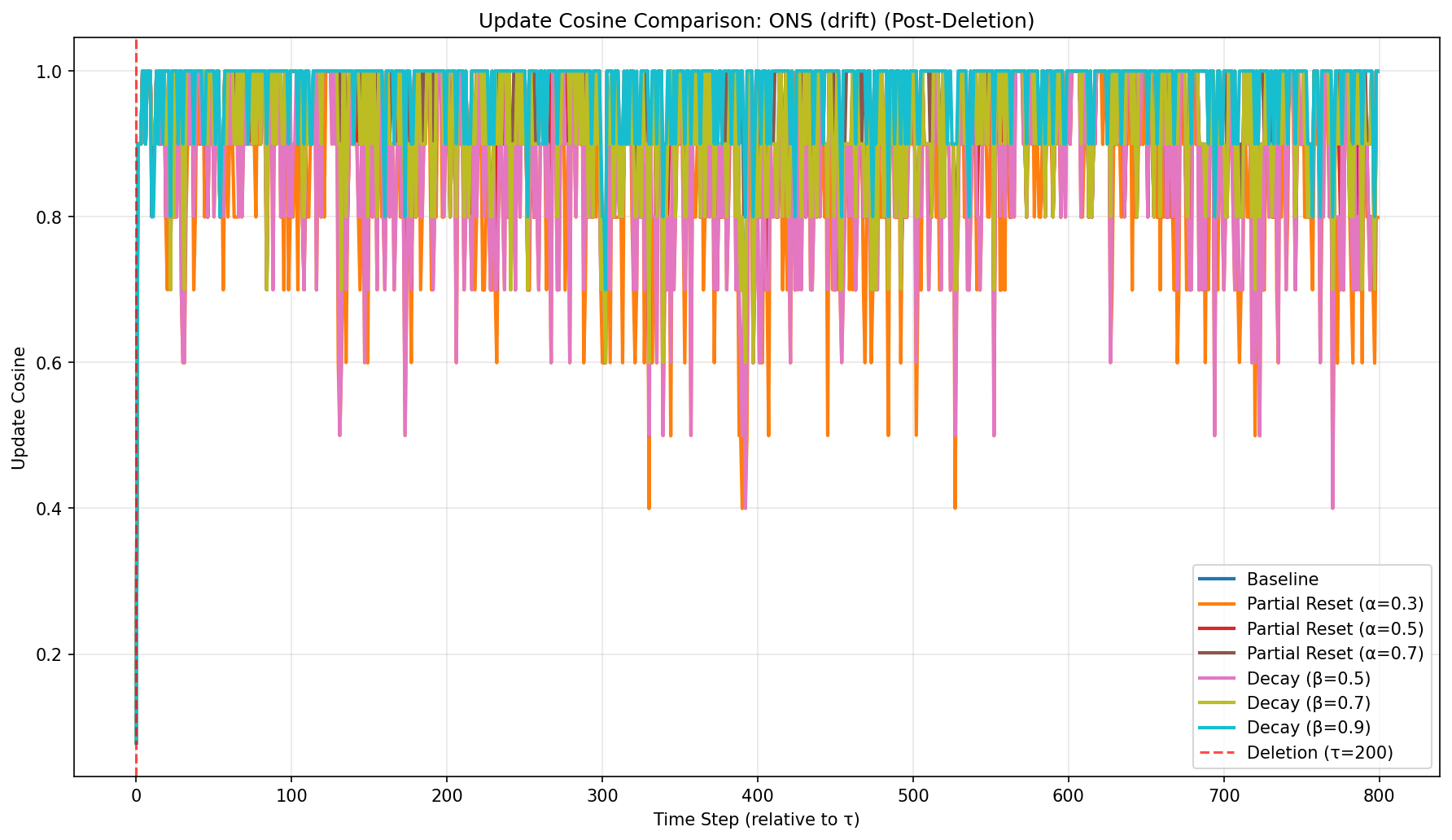}}{The figure shows the cosine similarity of the ONS optimizer with the ideal unlearned model. The cosine similarity indicates the difference between the observed and counterfactual optimizer states. As soon as deletion occurs, the observed models experience cosine volatility proportional to the strength of the intervention. The models with the lightest interventions ($\beta = 0.3$, $\alpha = 0.5$) show the heaviest volatility in terms of cosine dissimilarity, indicating some amount of residual information. The comparators showing the least dissimilarity are those experiencing near-total eigendecay as part of the deletion ($\beta = 0.9$, $\alpha = 0.7$)}
  \caption{Spectral information was recorded at every step to validate the treatments and track the difference in spectra between the observed and counterfactual iterates.}
  \label{fig:spectral-decomposition}
\end{figurepair}

The trajectory-level plots provide the missing information that regret alone cannot capture. Figure~\ref{fig:tracking-errors} shows that tracking error remains low in absolute terms but does not converge to the counterfactual path after deletion. Instead, the treatments separate after $\tau$, with stronger interventions generally producing larger downstream deviation. The important point is not that performance fails outright, but that the learner continues along a post-deletion trajectory that is not the one it would have followed had the deleted data never been observed.

The spectral diagnostics in Figure~\ref{fig:spectral-decomposition} sharpen this conclusion. The trace of $A_t$ continues to evolve smoothly after deletion, indicating that learning proceeds normally at the level of aggregate gradient accumulation. By contrast, the condition number and cosine-based alignment diagnostics exhibit discontinuities and increased volatility after intervention. This combination implies that deletion perturbs the shape and orientation of the second-order state more than it perturbs the overall scale of learning.

These resultes therefore suggest that deletion does not primarily manifest as persistent first-order failure, but as persistent geometric hysteresis. The learner remains effective in terms of regret, yet its internal state is no longer aligned with the counterfactual optimizer that never processed the deleted observations.

\section{Discussion}
We test the hypothesis that unlearned information may persist in the state of second-order optimizers and find substantial evidence in support of the claim. While second-order methods are popular for their quick adaptation to nonstationarity, this memory comes with deep structural knowledge encoded into the state's geometry. Any unlearning intervention needs to modify the entire state: parameters and aggregate.

The experiments show that unlearning difficulty in second-order methods is not primarily a performance issue but a state alignment problem. Reducing stored geometry increases volatility without improving model inference, suggesting a class of equivalently sufficient solutions in parameter space.

This suggests a fundamental trade-off: richer optimizer state enables faster adaptation but increases path dependence under deletions. Effective unlearning in such systems requires principled control over optimizer memory, not merely parameter updates.

\section{Future Work}

We compare the machine unlearning problem to that of a misaligned state in two tightly-coupled aggregates. As such, there is room to abstract the notion of unlearning from second-order aggregates to higher-order information states. This moves the problem from one of parameter similarity in the first-order to one of state alignment in higher-order information states. This would provide a rigorous theoretical backbone to unlearning methods in Quasi-Newton optimizers, including online L-BFGS and Hessian Vector Products, and move the problem firmly into one of Markov Decision Processes and regret control policies.

Similarly, there is insufficient work of certified unlearning in an online setting. Existing methods require large amount of compute and memory for step size calculations, restricting their use in IoT and memory-constrained environments. An opportunity exists to define an algorithm that provides guarantees for learning while guaranteeing tight regret bounds and an aligned information state.

\bibliographystyle{plain}
\bibliography{unlearning-geometry}
\end{document}